\documentclass[letterpaper, 10 pt, conference]{ieeeconf}  

\IEEEoverridecommandlockouts                              

\usepackage{color}
\usepackage{amssymb}
\usepackage{amsmath}
\usepackage{graphicx}
\usepackage{amsmath}
\usepackage{subfig}
\usepackage{graphicx}
\usepackage{amsfonts}
\usepackage{bm}
\usepackage{wrapfig}
\usepackage{makeidx}
\usepackage{moreverb}
\usepackage[lined,boxed,commentsnumbered]{algorithm2e}
\usepackage{multirow}
\usepackage{array}
\usepackage{MnSymbol}
\usepackage{bbding}
\usepackage{booktabs,tabularx}

\newcommand{\degree}{\ensuremath{^\circ}}
\DeclareMathOperator{\Tr}{Tr}

\DeclareGraphicsExtensions{.jpg,.png,.tif}

\title{A Hierarchical Approach for Joint Multi-view Object Pose Estimation and Categorization}

\author{Mete Ozay$^{1}$, Krzysztof Walas$^{1,2}$ and Ale\v{s} Leonardis$^{1}$
\thanks{This work was supported in part by the European Commission project PaCMan EU FP7-ICT, 600918.}
\thanks{$^{1}$Mete Ozay, Krzysztof Walas and Ale\v{s} Leonardis are with School of Computer Science,
        University of Birmingham, Edgbaston B15 2TT Birmingham, United Kingdom
        {\tt\small \{m.ozay,walask,a.Leonardis\}@cs.bham.ac.uk}}%
\thanks{$^{2}$Krzysztof Walas is also with Department of Electrical Engineering, Poznan University of Technology, ul. Piotrowo 3a, 60-965 Poznan, Poland
        {\tt\small krzysztof.walas@put.poznan.pl}}%
}

\begin{document}

\maketitle
\thispagestyle{empty}
\pagestyle{empty}

\begin{abstract}
We propose a joint object pose estimation and categorization approach which extracts information about object poses and categories from the object parts and compositions constructed at different layers of a hierarchical object representation algorithm, namely Learned Hierarchy of Parts (LHOP) \cite{fidler_cvpr07}. In the proposed approach, we first employ the LHOP to learn hierarchical part libraries which represent entity parts and compositions across different object categories and views. Then, we extract statistical and geometric features from the part realizations of the objects in the images in order to represent the information about object pose and category at each different layer of the hierarchy. Unlike the traditional approaches which consider specific layers of the hierarchies in order to extract information to perform specific tasks, we combine the information extracted at different layers to solve a joint object pose estimation and categorization problem using distributed optimization algorithms. We examine the proposed generative-discriminative learning approach and the algorithms on two benchmark 2-D multi-view image datasets. The proposed approach and the algorithms outperform state-of-the-art classification, regression and feature extraction algorithms. In addition, the experimental results shed light on the relationship between object categorization, pose estimation and the part realizations observed at different layers of the hierarchy.
\end{abstract}

\section{Introduction}
The field of service robots aims to provide robots with functionalities which allow them to work in man-made environments. For instance, the robots should be able to categorize objects and estimate the pose of the objects to accomplish various robotics tasks, such as grasping objects \cite{Kootstra_ijrr12}. Representation of object categories enables the robot to further refine the grasping strategy by giving context to the search for the pose of the object \cite{lai_aaai11}. 

In this paper, we propose a joint object categorization and pose estimation approach which extract information about statistical and geometric properties of object poses and categories extracted from the object parts and compositions that are constructed at different layers of the Learned Hierarchy of Parts (LHOP) \cite{fidler_cvpr07,fidler_book,fidler_eccv10}. 

\begin{figure}[ht!]
      \centering
	  {\includegraphics[width=4in, height=2.45in]{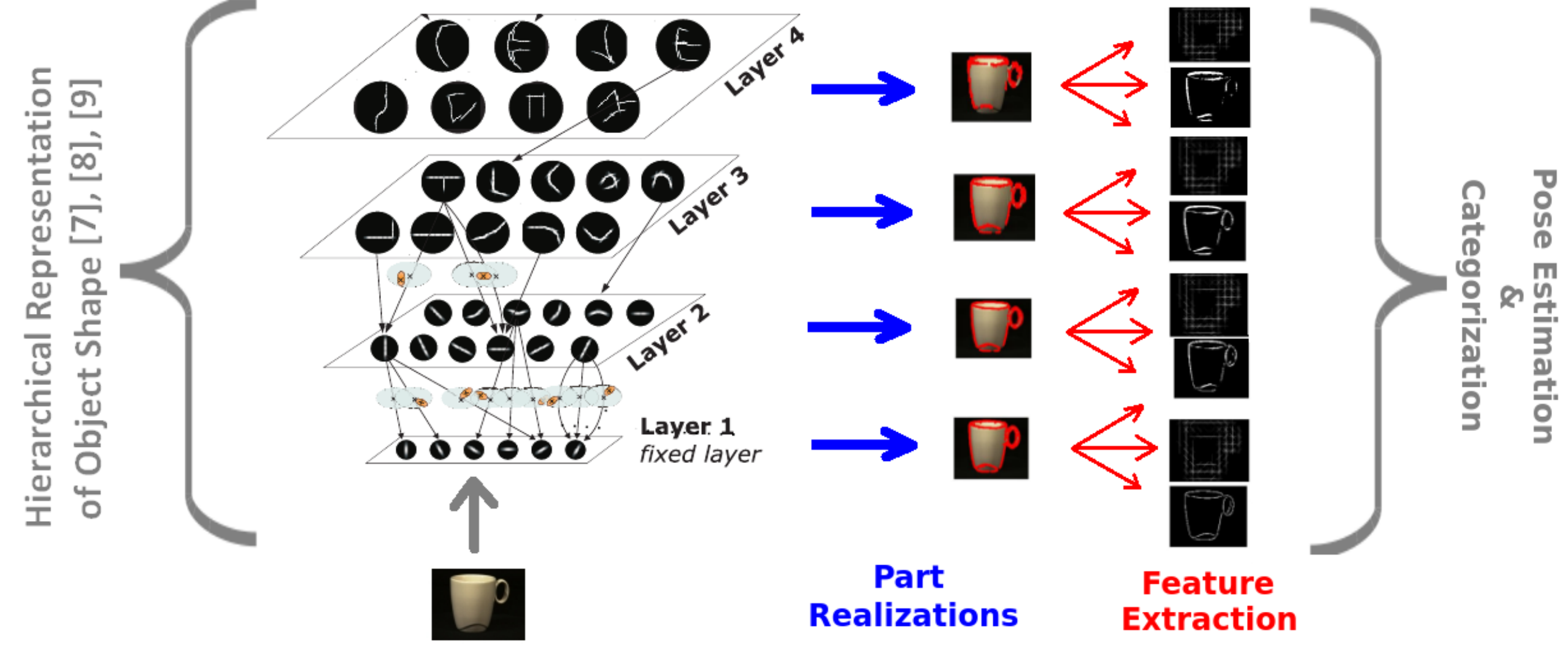}} 
      \caption{Combination of features extracted from part realizations detected at different layers of LHOP.}
   \label{fig:intro}
   \end{figure}
   
In the proposed approach, we first employ LHOP \cite{fidler_cvpr07,fidler_book} to learn hierarchical part libraries which represent object parts and compositions across different object categories and views as shown in Fig. \ref{fig:intro}. Then, we extract statistical and geometric features from the part realizations of the objects in the images in order to represent the information about the object pose and category at each different layer of the hierarchy. We propose two novel feature extraction algorithms, namely Histogram of Oriented Parts (HOP) and Entropy of Part Graphs. HOP features measure  local distributions of global orientations of part realizations of objects at different layers of a hierarchy. On the other hand, Entropy of Part Graphs provides information about the statistical and geometric structure of object representations by measuring the entropy of the relative orientations of parts. In addition, we compute a Histogram of Oriented Gradients (HOG) \cite{HOG} of part realizations in order to obtain information about the co-occurrence of the gradients of part orientations.

Unlike traditional approaches which extract information from the object representations at specific layers of the hierarchy to accomplish specific tasks, we combine the information extracted at different layers to solve a joint object pose estimation and categorization problem using a distributed optimization algorithm. For this purpose, we first formulate the joint object pose estimation and categorization problem as a sparse optimization problem called Group Lasso \cite{glasso}. We consider the pose estimation problem as a sparse regression problem and the object categorization problem as a multi-class logistic regression problem using Group Lasso. Then, we solve the optimization problems using a distributed and parallel optimization algorithm called the Alternating Direction Method of Multipliers (ADMM) \cite{admm}.

In this work, we extract information on object poses and categories from 2-D images to handle the cases where 3-D sensing may not be available or may be unreliable (e.g. glass, metal objects). We examine the proposed approach and the algorithms on two benchmark 2-D multiple-view image datasets. The proposed approach and the algorithms outperform state-of-the-art Support Vector Machine and Regression algorithms. In addition, the experimental results shed light on the relationship between object categorization, pose estimation and the part realizations observed at different layers of the hierarchy. 

In the next section, related work is reviewed and the novelty of our proposed approach is summarized. In Section \ref{sec:lhop}, a brief presentation of the hierarchical compositional representation is given. Feature extraction algorithms are introduced in Section \ref{sec:fe}. The joint object pose estimation and categorization problem is defined, and two algorithms are proposed to solve the optimization problem in Section \ref{sec:comb}. Experimental analyses are given in Section \ref{sec:exp}. Section \ref{sec:conc} concludes the paper.

\subsection{Related Work and  Contribution}

In the field of computer vision the problem of object categorization and pose estimation is studied thoroughly and some of the approaches are proliferating to the robotics community. With an advent of devices based on PrimeSense sensors, uni-modal 3-D or multi-modal integration of 2-D and 3-D data (e.g. rgb-d data) have been widely used by robotics researchers \cite{jiang_corr12}. However, 3-D sensing may not be available or reliable due to limitations of object structures, lighting resources and imaging conditions in many cases where single or multiple view 2-D images are used for categorization and pose estimation \cite{collet_icra09,collet_ijrr11,Damien}. In \cite{Damien}, a probabilistic approach is proposed to estimate the pose of a known object using a single image. Collet et al. \cite{collet_icra09} build 3D models of objects using SIFT features extracted from 2D images for robotic manipulation, and combine single image and multiple image object recognition and pose estimation algorithms in a framework in \cite{collet_ijrr11}. 


A promising approach to the object categorization and the scene description is the use of hierarchical compositional architectures \cite{fidler_cvpr07,fidler_eccv10,lai_aaai11}. Compositional hierarchical models are constructed for object categorization and detection using single images in \cite{fidler_cvpr07,fidler_eccv10}. Multiple view images are used for pose estimation and categorization using a hierarchical architecture in \cite{lai_aaai11}. In the aforementioned approaches, the tasks are performed using either discriminative or generative top-down or bottom-up learning approaches in architectures. For instance, Lai et al. employ a top-down categorization and pose estimation approach in \cite{lai_aaai11}, where a different task is performed at each different layer of the hierarchy. Note that, a categorization error occurring at the top-layer of the hierarchy may propagate to the lower layer and affect the performance of other tasks such as pose estimation in this approach. In our proposed approach, we first construct generative representations of object shapes using LHOP \cite{fidler_cvpr07,fidler_book,fidler_eccv10}. Then, we train discriminative models by extracting features from the object representations. In addition, we propose a new method, which enables us to combine the information extracted at each different layer of the hierarchy, for joint categorization and pose estimation of objects. We avoid the propagation of errors of performing multiple tasks through the layers and enable the shareability of parts among layers by the employment of optimization algorithms in each layer in a parallel and distributed learning framework.

The novelty of the proposed approach and the paper can be summarized as follows;

\begin{enumerate}

\item In this work, the Learned Hierarchy of Parts (LHOP) is employed in order to learn a hierarchy of parts using the shareability of parts across different views as well as different categories \cite{fidler_cvpr07,fidler_book}. 
\item Two novel feature extraction algorithms, namely Histogram of Oriented Parts (HOP) and Entropy of Part Graphs, are proposed in order to obtain information about the statistical and geometric structure of objects' shapes represented at different layers of the hierarchy using part realizations.
\item  The proposed generative-discriminative approach enables us to combine the information extracted at different layers in order to solve a joint object pose estimation and categorization problem using a distributed and parallel optimization algorithm. Therefore, this approach also enables us to share the parts among different layers and avoid the propagation of object categorization and pose estimation errors through the layers.

\end{enumerate}

\section{Learned Hierarchy of Parts}
\label{sec:lhop}
In this section, Learned Hierarchy of Parts (LHOP)\cite{fidler_cvpr07,fidler_book} is briefly described. In LHOP, the object recognition process is performed in a hierarchy starting from a feature layer through more complex and abstract interpretations of object shapes to an object layer. A learned vocabulary is a recursive compositional representation of shape parts. Unsupervised bottom-up statistical learning is encompassed in order to obtain such a description. 

Shape representations are built upon a set of compositional parts which at the lowest layer use atomic features, e.g. Gabor features, extracted from image data. The object node is a composition of several child nodes located at one layer lower in the hierarchy, and the composition rule is recursively applied to each of its child nodes to the lowest layer $\Gamma_1$. All layers together form a hierarchically encoded vocabulary $\Gamma = \Gamma_1 \cup \Gamma_2 \cup \ldots \cup \Gamma_L$. The entire vocabulary $\Gamma$ is learned from the training set of images together with the vocabulary parameters \cite{fidler_book}. 

The parts in the hierarchy are defined recursively in the following way. Each part in the $l^{th}$ layer represents the spatial relations between its constituent subparts from the layer below. Each composite part $\mathcal{P}_k^l$ constructed at the $l^{th}$ layer is characterized by a central subpart $\mathcal{P}_{central}^{l-1}$ and a list of remaining subparts with their positions relative to the center as
\begin{equation}
 \mathcal{P}_k^l = (\mathcal{P}_{central}^{l-1},\{(\mathcal{P}_j^{l-1},\boldsymbol{\mu}_j,\Sigma_j)\}j),
\end{equation}
where $\boldsymbol{\mu}_j = (x_j,y_j)$ denotes the relative position of the subpart $\mathcal{P}_j^{l-1}$, while $\Sigma_j$ denotes the allowed variance of its position around ($x_j,y_j$).

\section{Feature Extraction from Learned Parts}
\label{sec:fe}

LHOP provides information about different properties of objects, such as poses, orientations and category memberships, at different layers \cite{fidler_cvpr07}. For instance, the information on shape parts, which are represented by edge structures and textural patterns observed in images, is obtained using Gabor features at the first layer $L_1$. In the second and the following layers, compositions of parts are constructed according to the co-occurrence of part realizations that are detected in the images among different views of the objects and across different object categories. In other words, a library of object parts and compositions is learned jointly for all object views and categories.  

In order to obtain information about statistical and geometric properties of parts, we extract three types of features from the part realizations detected at each different layer of the LHOP. 

\subsection{Histogram of Orientations of Parts}
\label{sec:hop}

\begin{figure}[thpb]
      \centering
	  {\includegraphics[width=2.8in, height=2.00in]{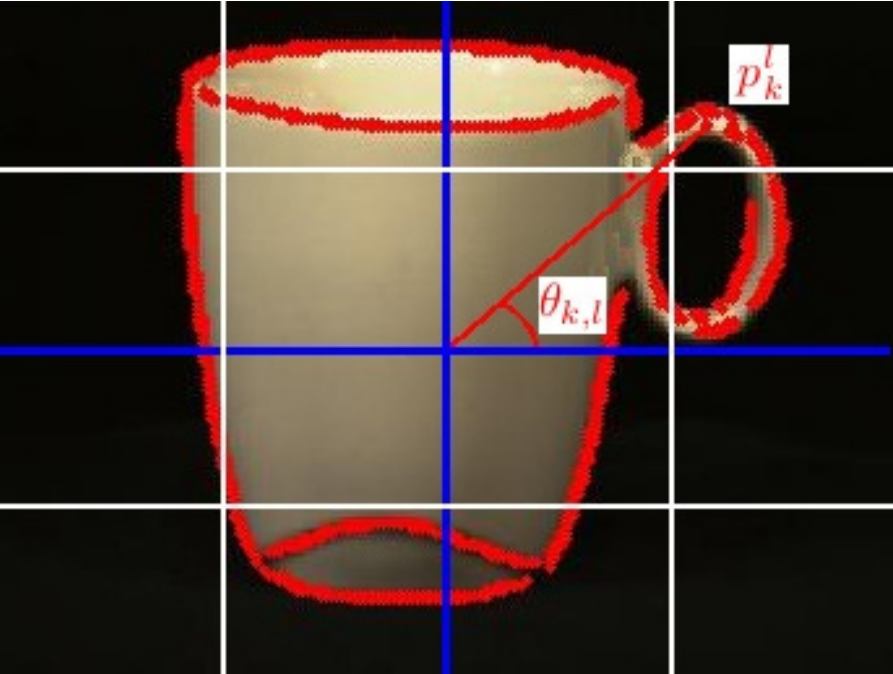}} 
      \caption{An image is partitioned into cells for the computation of histograms of orientations of parts. A part realization {\color{red} $p_k^l$} is depicted with a red point and associated to a part orientation degree {\color{red} $\theta_{k,l}$}.}

      \label{fig:hop}
   \end{figure}

Histograms of orientations of parts are computed in order to extract information on the co-occurrence of  orientations of the parts across different poses of objects. Part orientations are computed according to a coordinate system of an image $I$ whose origin is located at the center of the image $I$, and the axes of the coordinate system are shown with blue lines in Fig.~\ref{fig:hop}. 

If we define $p_k^l, \forall k=1,2, \ldots,K, \forall l=1,,2 \ldots,L$ as the realization of the $k^{th}$ detected part in the $l^{th}$ layer at an image coordinate $(x_k, y_k)$ of $I$, then its orientation with respect to the origin of the coordinate system is computed as
\[
\theta_{k,l}=\arctan(\frac{y_k}{x_k}).
\]
Then, the image $I$ is partitioned into $M$ cells $\{ I_m \}_{m=1}^M$, and histograms of the part orientations $\{ \theta_{k,l} \}_{k=1}^{K'}$ of the part realizations $\{ p_{k,l} \}_{k=1}^{K'}$ that are located in each cell $I_m$ are computed. The aggregated histogram values are considered as variables of a $D_p$ dimensional feature vector $\mathbf{f}_{hop}^l \in \mathbb{R}^{D_p}$.

\subsection{Histogram of Oriented Gradients of Parts}
\label{sec:hog}

In addition to the computation of histograms of orientations of part realizations $p_k^l,  \forall k=1,2, \ldots,K, \forall l=1,2, \ldots,L$, we compute histogram of oriented gradients (HOG) \cite{HOG} of $p_k^l$ in order to extract information about the distribution of gradient orientations of $p_k^l, \forall k,l$. We denote the HOG feature vector extracted using $\{ p_k^l \} _{k=1} ^K$ in the $l^{th}$ layer as $\mathbf{f}_{hog}^l \in \mathbb{R}^{D_h}$, where $D_h$ is the dimension of the HOG feature vector. The details of the implementation of HOG feature vectors are given in Section \ref{sec:exp}.

\subsection{The Entropy of Part Graphs}
\label{sec:vne}

We measure the statistical and structural properties of relative orientations of part realizations by measuring the complexity of a graph of parts. Mathematically speaking, we define a weighted undirected graph $G_l:=(E_l,V_l)$ in the $l^{th}$ layer, where $V_l:=\{ p_k^l \}$ is the set of part realizations, $E_l:=\{ e_{k',k} \}_{k',k=1} ^K$ is the set of edges, where each edge $e_{k',k}$ that connects the part realizations $p_{k'}^l$ and $p_k^l$ is associated to an edge weight $w_{k',k}$, which is defined as 
\[
w_{k',k}:=\arccos (\frac{\mathbf{pos}_{k'} \cdot \mathbf{pos}_{k}}{ \| \mathbf{pos}_{k'} \|_2 \| \mathbf{pos}_{k} \|_2 }),
\]
where $\mathbf{pos}_{k}:=(x_k,y_k)$ is the position vector of $p_{k'}^l$, $\| \cdot \|_2$ is the $\ell_2$ norm or Euclidean norm, and $\mathbf{pos}_{k'} \cdot \mathbf{pos}_{k}$ is the inner product of $\mathbf{pos}_{k'}$ and $\mathbf{pos}_{k}$. In other words, the edge weights are computed according to the orientations of parts relative to each other. 

We measure the complexity of the weighted graph by computing its graph entropy. First, we compute the normalized weighted graph Laplacian $\mathcal{L}$ \cite{von,ent} as 
\[
\mathcal{L}= \frac{1}{K(K-1)} (\mathcal{D}-\mathcal{W}),
\] 
where $\mathcal{W} \in \mathbb{R}^{K \times K}$ is a weighted adjacency matrix or a matrix of weights $w _{k',k}$, and $\mathcal{D} \in \mathbb{R}^{K \times K}$ is a diagonal matrix with members $\mathcal{D}_{k,k} := \sum \limits _{k'=1} ^K w _{k',k}$. Then, we compute the von Neumann entropy of $G_l$ \cite{von,ent} as
\begin{eqnarray}
S(G_l) &=&- \Tr (\mathcal{L} \log_2 \mathcal{L}) \\
&=& -\sum \limits _{k=1} ^K \nu_k,
\end{eqnarray}
where $\nu_1 \geq \nu_2 \geq \ldots \geq \nu_k \geq \ldots \geq \nu_K=0$ are the eigenvalues of $\mathcal{L}$, $\Tr( \mathcal{L} \log_2 \mathcal{L})$ is the trace of the matrix product $\mathcal{L} \log_2 \mathcal{L}$ and $0 \log_2 0=0$. We use $S(G_l)$ as a feature variable $f_{ent}^l:=S(G_l)$.

\section{Combination of Information Obtained at Different Layers of LHOP for Joint Object Pose Estimation and Categorization}
\label{sec:comb}

In hierarchical compositional architectures, a different object property, such as object shape, pose and category, is represented at a different layer of a hierarchy in a vocabulary \cite{lai_aaai11}. According the structures of the abstract representations of the properties, i.e. vocabularies, recognition processes have been performed using either a bottom-up \cite{fidler_cvpr07,fidler_book} or top-down \cite{lai_aaai11} approach. It's worth noting that the information in the representations are distributed among the layers in the vocabularies. In other words, the information about the category of an object may reside at the lower layers of the hierarchy instead of the top layer. In addition, lower layer atomic features, e.g. oriented Gabor features, provide information about part orientations which can be used for the estimation of pose and view-points of objects at the higher layers. Moreover, the relationship between the pose and category of an object is bi-directional. Therefore, an information integration approach should be considered in order to avoid the propagation of errors that occur in multi-task learning and recognition problems such as joint object categorization and pose estimation, especially when only one of the bottom-up and top-down approaches is implemented.

For this purpose, we propose a generative-discriminative learning approach in order to combine the information obtained at each different layer of LHOP using the features extracted from part realizations. We represent the features defining a $D_p+D_h+1$ dimensional feature vector $\mathbf{f}^l=(\mathbf{f}_{hop}^l,\mathbf{f}_{hog}^l,f_{ent}^l)$. The feature vector $\mathbf{f}^l$ is computed for each training and test image, therefore we denote the feature vector of the $i^{th}$ image $I_i$ as $\mathbf{f}_i^l$, $\forall i=1,2,\ldots,N$, in the rest of the paper.

We combine the feature vectors extracted at each $l^{th}$ layer for object pose estimation and categorization under the following Group Lasso optimization problem \cite{glasso}
\begin{eqnarray}
\text{minimize} & \| \mathcal{F} \boldsymbol{\omega} - \mathbf{z} \| ^2 _2 + \lambda \displaystyle \sum \limits ^L _{l=1} \| \boldsymbol{{\omega}}_l \| _2 ,
\label{eq:glasso}
\end{eqnarray}
where $\| \cdot \|_2 ^2$ is the squared $\ell_2$ norm, $\lambda \in \mathbb{R}$ is a regularization parameter, $\boldsymbol{{\omega}}_l$ is the weight vector computed at the $l^{th}$ layer, $\mathcal{F} \in \mathbb{R}^{N \times L}$ is a matrix of feature vectors $\mathbf{f}_i^l$, $\forall i=1,2,\ldots,N$, $\forall l=1,2,\ldots,L$ and $\mathbf{z} = (z_1,z_2,\ldots,z_N)$ is a vector of target variables $z_i \in \mathbb{R}$, $\forall i=1,2,\ldots,N$. More specifically, $z_i \in \Omega$ where $\Omega$ is a set of object poses, i.e. object orientation degrees, in a pose estimation problem.

We solve \eqref{eq:glasso} using a distributed optimization algorithm called Alternating Direction Method of Multipliers \cite{admm}. For this purpose, we first re-write \eqref{eq:glasso} in the ADMM form as follows
\begin{eqnarray}
\begin{matrix} 
& \text{minimize} && \| \mathcal{F} \boldsymbol{\phi} - \mathbf{z} \| ^2 _2 + \lambda \displaystyle \sum \limits ^L _{l=1} \| \boldsymbol{{\omega}}_l \| _2 \\
& \text{subject to} && \boldsymbol{ {\omega}}_l - \boldsymbol{\hat {\phi}}_l =\mathbf{0} \;, l=1,2,\ldots,L \; ,
\end{matrix} 
\label{eq:admm}
\end{eqnarray}
where $\boldsymbol{\hat {\phi}}_l$ is the local estimate of the global variable $\boldsymbol{ {\phi}}$ for $\boldsymbol{ {\omega}}_l $ at the $l^{th}$ layer. Then, we solve \eqref{eq:admm} in the following three steps \cite{admm,jsac},

\begin{enumerate}
\item At each layer $l$, we compute $\boldsymbol{{\omega}}_l^{t+1}$ as
\begin{equation}
\boldsymbol{{\omega}}_l ^{t+1} :=  \underset{\boldsymbol{{\omega}}_l}{\operatorname{argmin}} \Big ( \rho \| \boldsymbol{\mu}_l^t  \| _2^2 + \lambda \| \boldsymbol{{\omega}}_l \| _2 \Big ), 
\label{eq:w_update1}
\end{equation}
where $\boldsymbol{\mu}_l^t=\mathcal{F}_l(\boldsymbol{{\omega}}_l-\boldsymbol{{\omega}}_l ^{t}) - \boldsymbol{\bar {\phi}} ^t + \mathbf{a}^t + \overline{\mathcal{F}_l\boldsymbol{\omega}_l}^t$, $\rho >0$ is a penalty parameter, $\overline{\mathcal{F}_l\boldsymbol{\omega}_l}^t=\frac{1}{L} \sum\limits _{l=1}^L \mathcal{F}_l\boldsymbol{\omega}_l^t$, $\boldsymbol{\bar {\phi}} ^t$ is the average of $\boldsymbol{ {\phi}}_l ^t$, $\forall l=1,\ldots,L$, and $\mathbf{a}^t$ is a vector of scaled dual optimization variables computed at an iteration $t$. 

\item Then we update $\boldsymbol{\hat {\phi}}_l$ as
\begin{equation}
\boldsymbol{\hat {\phi}}_l  ^{t+1} :=  \frac{1}{L+\rho} \Big( \mathbf{z} + \rho \overline{\mathcal{F}_l\boldsymbol{\omega}_l}^{t+1} + \rho \mathbf{a}^t  \Big ).
\label{eq:phi_update1}
\end{equation}

\item Finally, $\mathbf{a}$ is updated as
\begin{equation}
\mathbf{a} ^{t+1} :=  \mathbf{a}^t + \overline{\mathcal{F}_l\boldsymbol{\omega}_l}^t - \boldsymbol{\hat {\phi}}_l  ^{t+1}.
\label{eq:a_update1}
\end{equation}
\end{enumerate}
These three steps are iterated until a halting criterion, such as $t \geq T$ for a given termination time $T$, is achieved. Implementation details are given in the next section.

In a $C$ class object categorization problem, $z_i \in \{1,2, \ldots, c, \ldots,C \}$ is a category variable. In order to solve this problem, we employ \textit{1-of-C} coding for sparse logistic regression as

\begin{eqnarray}
P(z_i^c = 1 | \mathbf{f}_i)&=& \frac{\exp(h_j(\mathbf{f}_i))}{1+ \exp(h_c(\mathbf{f}_i))},
\label{eq:log_reg1}
\end{eqnarray}
where $h_c(\mathbf{f}_i)= \mathbf{f}_i \cdot \boldsymbol{{\omega}^c}$, $\boldsymbol{{\omega}^c}$ is a weight vector associated to the $c^{th}$ category, $z_i^c = 1$ if $z_i = c$, $\forall i=1,2,\ldots,N$. Then, we define the following optimization problem 

\begin{eqnarray}
\text{minimize} & -\sum \limits _{l=1} ^L \sum \limits _{i=1} ^N loss_l(i) + \lambda \| \boldsymbol{{\omega^c}} \| _1,
\label{eq:opt_logreg}
\end{eqnarray}
where $loss_l(i)=z_i^c h_c(\mathbf{f}_i)- \log \Big( \exp(h_c(\mathbf{f}_i)) +1 \Big)$. In order to solve \eqref{eq:opt_logreg}, we employ the three update steps given above with two modifications. First, we solve \eqref{eq:w_update1} for the $\ell_1$ norm in the last regularization term $\lambda \| \boldsymbol{{\omega}}_l \| _1$ instead of the $\ell_2$ norm. Second, we employ the logistic regression loss function in the computation of $\boldsymbol{\hat {\phi}}_l$ as
\begin{equation}
\boldsymbol{\hat {\phi}}_l  ^{t+1} :=  \underset{\boldsymbol{{\phi}}_l}{\operatorname{argmin}} \Big ( \rho \| \boldsymbol{ {\phi}_l}-\overline{\mathcal{F}_l\boldsymbol{\omega}_l}^{t+1}-\mathbf{a}^t \| _2 
+\log(1+\exp-(L\boldsymbol{{\phi}}_l))
\Big ).
\label{eq:phi_update2}
\end{equation}

In the training phase of the pose estimation algorithm, we compute the solution vector $\boldsymbol{{\omega}} = (\boldsymbol{{\omega}}_1,\boldsymbol{{\omega}}_2, \ldots,\boldsymbol{{\omega}}_L  \}$ using training data. In the test phase, we employ the solution vector $\boldsymbol{{\omega}}$ on a given test feature vector $\mathbf{f}_i$ of the part realizations of an object to estimate its pose as 
\[
\hat{z}_i=\mathbf{f}_i \cdot \boldsymbol{\omega}.
\]

In the categorization problem, we predict the category label $\hat{z}_i$ of an object in the $i^{th}$ image as 
\[
\hat{z}_i=\underset{c}{\operatorname{argmax}} \; \hat{z}_i^c.
\]

\section{Experiments}
\label{sec:exp}

We examine our proposed approach and algorithms on two benchmark object categorization and pose estimation datasets, which are namely the Amsterdam Library of Object Images (ALOI) \cite{aloi} and the Columbia Object Image Library (COIL-100) \cite{coil}. We have chosen these two benchmark datasets for two main reasons. First, images of objects are captured by rotating the objects on a turntable by regular orientation degrees which enable us to analyze our proposed algorithm for multi-view object pose estimation and categorization in uncluttered scenes. Second, object poses and categories are labeled within \textit{acceptable} precision which is important to satisfy the statistical stability of training and test samples and their target values. In our experiments, we also re-calibrated labels of pose and rotation values of the objects that are mis-recorded in the datasets. 

We select the bin size ($bSize$) of the histograms and cell size $M$ of HOP (see Section \ref{sec:hop}) and HOG features (see Section \ref{sec:hog}) by greedy search on the parameter set $\{ 8,16,32,64 \}$, and take the \textit{optimal} $\hat{bSize}$ and $\hat{M}$ which minimizes pose estimation and categorization errors in pose estimation and categorization problems using training datasets, respectively. In the employment of optimization algorithms, we compute $\lambda=\alpha\lambda_{\max}$, where  $\lambda_{\max}= \|\mathcal{F} \boldsymbol{\omega} \|_{\infty}$,  $\boldsymbol{\omega}=(\boldsymbol{\omega}_1,\ldots,\boldsymbol{\omega}_L)$, $\| \cdot \|_{\infty}$ is $\ell_{\infty}$ norm and $\alpha$ parameter is selected from the set $\{ 10^{-6},10^{-5},\ldots, 10^{1} \}$ using greedy search by minimizing training error of object pose estimation and categorization as suggested in \cite{admm}. In the implementation of LHOP, we learn the compositional hierarchy of parts and compute the part realizations for $L=1,2,3,4$ \cite{fidler_cvpr07}.

In the experiments, pose estimation and categorization performances of the proposed algorithms are compared with state-of-the-art Support Vector Regression (SVR), Support Vector Machines (SVM) \cite{libsvm}, Lasso and Logistic regression algorithms \cite{stl} which use the state-of-the-art HOG features \cite{HOG} extracted from the images as considered in \cite{pose1}. In the results, we refer to an implementation of SVM with HOG features as SVM-HOG, SVM with the proposed LHOP features as SVM-LHOP, SVR with HOG features as SVR-HOG, SVR with the proposed LHOP features as SVR-LHOP, Lasso with HOG features as L-HOG, Logistic Regression with HOG features as LR-HOG, Lasso with LHOP features as L-LHOP, Logistic Regression with LHOP features as LR-LHOP. 

We use RBF kernels in SVR and SVM. The kernel width parameter $\sigma$ is searched in the interval $\log(\sigma) \in [-10,5]$  and the SVR cost penalization parameter $\epsilon$ is searched in the interval $\log(\epsilon) \in [-10,5]$ using the training datasets.

\subsection{Experiments on Object Pose Estimation}
\label{sec:exp_pe}

We have conducted two types of experiments for object pose estimation, namely \textit{Object-wise} and \textit{Category-wise} Pose Estimation. We analyze the sharability of the parts across different views of an object in Object-wise Pose Estimation experiments. In Category-wise Pose Estimation experiments, we analyze incorporation of category information to sharability of parts in the LHOP and to pose estimation performance.

\subsubsection{Experiments on Object-wise Pose Estimation}
\label{sec:exp_ow}

In the first set of experiments, we consider the objects belonging to each different category, individually. For instance, we select $\aleph_{tr} ^o=4$ objects for training and $\aleph_{te} ^o=1$ objects for testing using objects belonging to \textbf{cups} category. The ID numbers of the objects and their category names are given in Table~\ref{tab:table1}. For each object, we have $72$ object instances each of which represents an orientation of the object $z_i=\Theta_i$ on a turntable rotated with $\Theta_i \in \Omega$ and $\Omega= \{ 0\degree, 5\degree, 10\degree, \ldots,355\degree \}$.

\begin{table*}[ht!]
  \centering
  \caption{The samples that are selected from ALOI dataset and used in Object-wise Pose Estimation Experiments}
  \centering
    \begin{tabular}{|>{\centering\arraybackslash}m{1.5cm}|>{\centering\arraybackslash}m{1.1cm}|>{\centering\arraybackslash}m{1.1cm}|>{\centering\arraybackslash}m{1.1cm}|>{\centering\arraybackslash}m{1.1cm}|>{\centering\arraybackslash}m{1.1cm}|>{\centering\arraybackslash}m{1.1cm}|>{\centering\arraybackslash}m{1.1cm}|>{\centering\arraybackslash}m{1.1cm}|}
   \hline
     {\textbf{Category Name}} & {\textbf{Apples}} & {\textbf{Balls}} & {\textbf{Bottles}} & {\textbf{Boxes}} & \textbf{Cars} & {\textbf{Cups}} & {\textbf{Shoes}}\\
     \hline
     \textbf{Object IDs for Training} & 82 & 103 & 762 & 13& 54 & 157 & 9 \\
     \hline   
     \textbf{Object IDs for Testing} & 363, 540, 649, 710  & 164, 266, 291, 585 & 798, 829, 831, 965 & 110, 26, 46, 78 & 136, 138, 148, 158 & 36, 125, 153, 259 & 93, 113, 350, 826\\
     \hline
    \end{tabular}%
  \label{tab:table1}%
\end{table*}%

\begin{figure}[ht!]
      \centering
	  {\includegraphics[width=3.5in, height=4.5in]{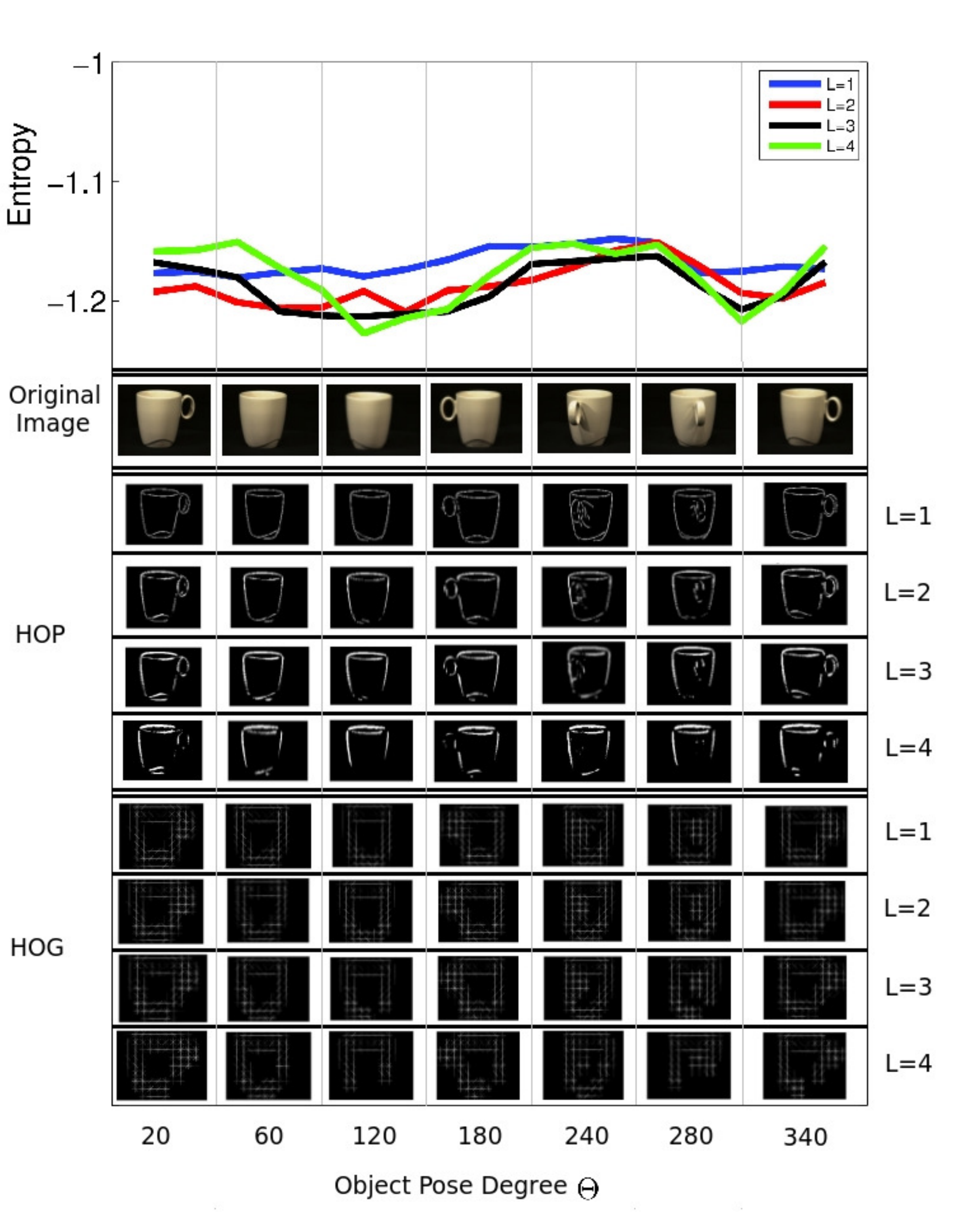}} 
      \caption{Visualization of features extracted from part realizations for each different orientation of a cup and at each different layer of LHOP.}

      \label{fig:all}
   \end{figure}

In the experiments, we first analyze the variation of part realizations and feature vectors across different orientations of an object. We visualize the features $\mathbf{f}_{hop}^l$, $\mathbf{f}_{hog}^l$ and $f_{ent}^l$ in Fig.~\ref{fig:all} for a cup which is oriented with $\Theta \in \{ 20\degree, 60\degree, 120\degree, 180\degree, 240\degree, 280\degree, 340\degree \}$  and for each $l=1,2,3,4$. In the first row at the top of the figure, the change of $f_{ent}^l$ is visualized $\forall l$. In the second row, the original images of the objects are given. In the third to the sixth rows, $\mathbf{f}_{hop}^l$ are visualized by displaying the part realizations with pixel intensity values $\| \mathbf{f}_{hop}^l \|_2 ^2$ for each $l=1,2,3,4$. $\mathbf{f}_{hog}^l$ features are visualized in the rest of the rows for each $l$. 

In Fig.~\ref{fig:all}, we first observe that $f_{ent}^{l=1}$ values of the object change discriminatively across different object orientations $\Theta$. For instance, if the handle of the cup is not seen from the front viewpoint of the cup (e.g. at $\Theta=60\degree,120\degree$), then we observe a smooth surface of the cup and the complexity of the part graphs, i.e. the entropy values, decrease. On the other hand, if the handle of the cup is observed at a front viewpoint (e.g. at $\Theta=240\degree,280\degree$), then the complexity increases. In addition, we observe that the difference between $f_{ent}^{l}$ values of the object parts across different orientations $\Theta$ decreases as $l$ increases. In other words, the discriminative power of the generative model of the LHOP increases at the higher layers of the LHOP since the LHOP captures the \textit{important} parts and compositions that are co-occurred across different views through different layers.

 \begin{figure}[htbp]
      \centering
	  {\includegraphics[width=4.5in, height=3.25in]{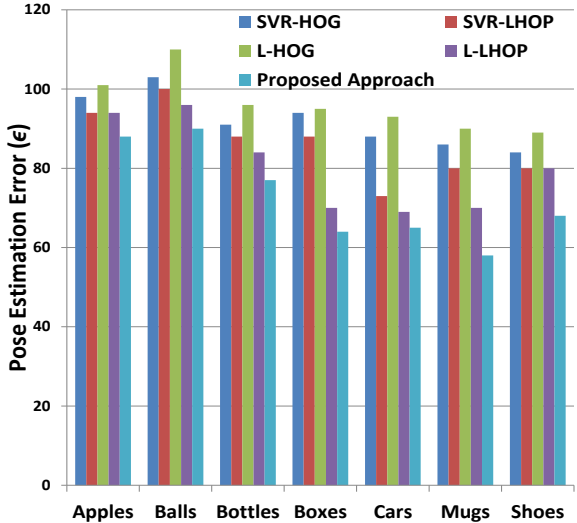}} 
      \caption{Comparison of Object-wise Pose estimation errors ($\epsilon$) of the proposed algorithms.}

      \label{fig:aloi_obj_pose}
   \end{figure}

Given a ground truth $\Theta$ and an estimated pose value $\hat{\Theta}$, the pose estimation error is defined as $\epsilon= || \Theta - \hat{\Theta}|| _2 ^2$. Pose estimation errors of state-of-the-art algorithms and the proposed Hierarchical Compositional Approach are given in Fig.~\ref{fig:aloi_obj_pose}. In these results, we observe that the pose estimation errors of the algorithms which are implemented using the symmetric objects, such as apples and balls, are greater than that of the algorithms that are implemented on more structural objects such as cups. 

 \begin{figure}[h!]
      \centering
	  {\includegraphics[width=3.5in, height=2.2in]{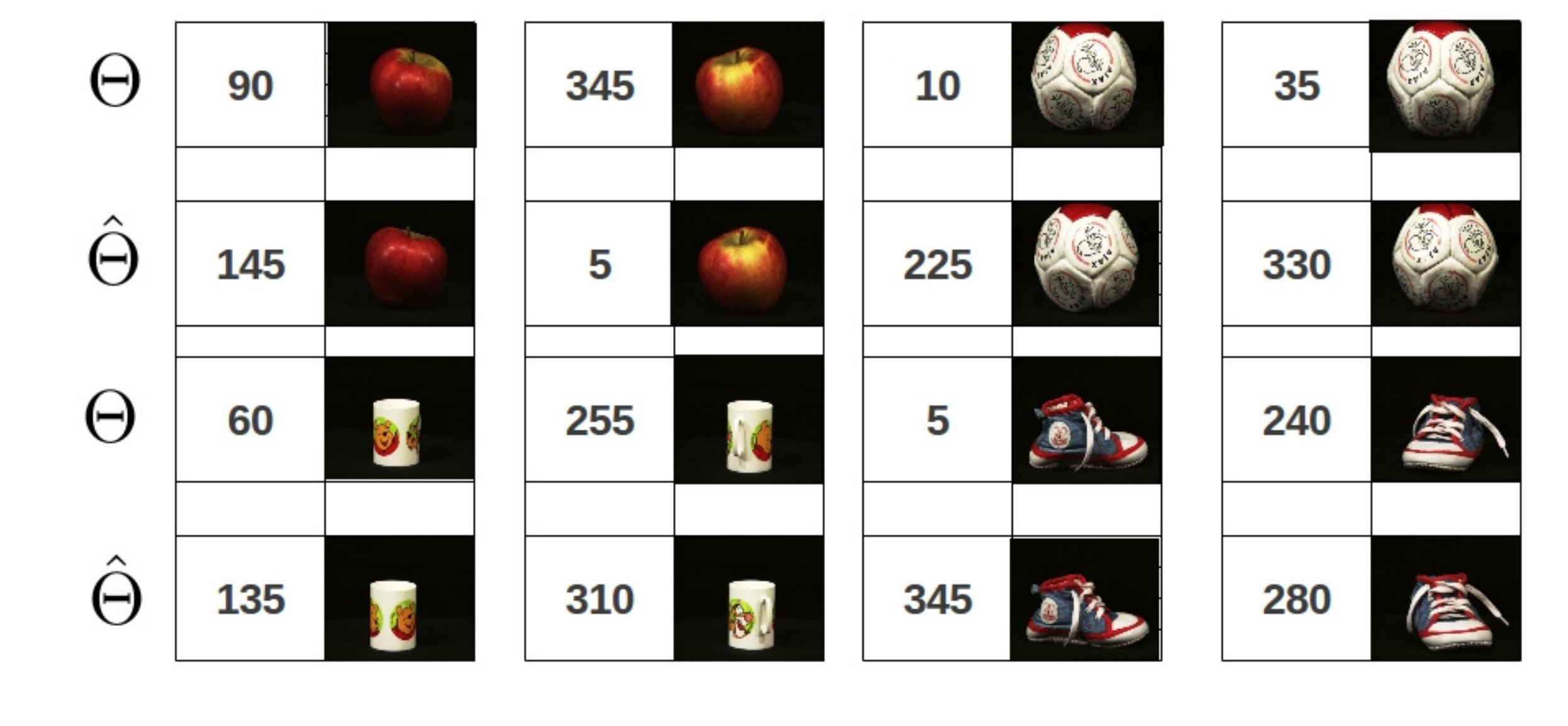}} 
      \caption{Results for some of the objects from Apples, Balls, Cups and Shoes categories obtained in Object-wise Pose estimation experiments.}
      \label{fig:est_orient}
   \end{figure}

In order to analyze this observation in detail, we show the ground truth $\Theta$ and the estimated orientations $\hat{\Theta}$ of some of the objects from Apples, Balls, cups and Shoes categories in Fig.~\ref{fig:est_orient}. We observe that some of the different views of the same object have the same shape and textural properties. For instance, the views of the ball at the orientations $\Theta=10\degree$ and $\Theta=225\degree$ represent the same pentagonal shape patterns. Therefore, similar parts are detected at these different views and the similar features are extracted from these detected parts. Then, the orientation of the ball, which is rotated by $\Theta=10\degree$, is incorrectly estimated as $\hat{\Theta}=225\degree$.

\subsubsection{Experiments on Category-wise Pose Estimation}
\label{sec:exp_cw}

In Category-wise Pose Estimation experiments, we select different $\aleph_{tr} ^o$ number of objects from different $C$ number of categories as training images to estimate the pose of test objects, randomly. We employ the experiments on both ALOI and COIL datasets. 

\begin{figure}[htbp]
      \centering
	  {\includegraphics[width=3.15in, height=0.8in]{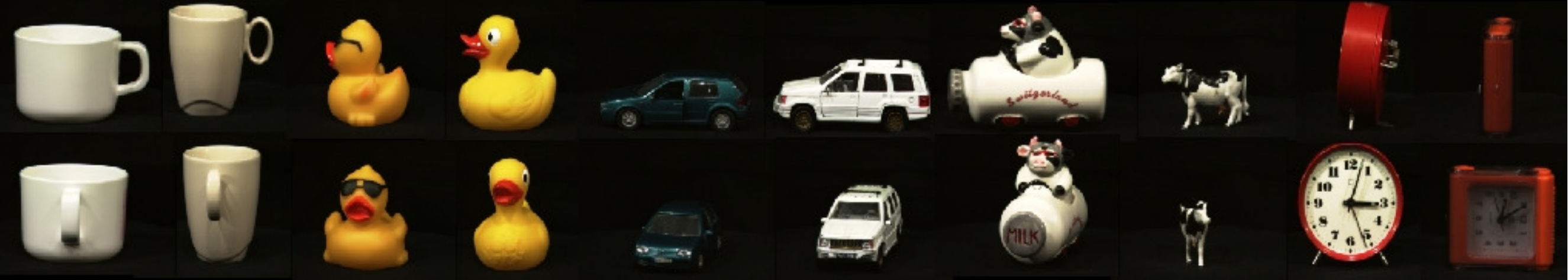}} 
      \caption{Sample images of the objects that are used in Category-wise Pose Estimation experiments. }

      \label{fig:aloi_examples}
   \end{figure}

In the ALOI dataset, we randomly select $\aleph_{tr} ^o=1,2,3,4$ number of training objects and $\aleph_{te} ^o=1$ test object which belong to Cups, Cow, Car, Clock and Duck categories. We repeat the random selection process two times and give the average pose estimation error for each experiment. In order to analyze the contribution of the information that can be obtained from the parts to the pose estimation performance using the part shareability of the LHOP, we initially select Cups and Cow categories ($C=2$) and add new categories (Car, Clock and Duck) to the dataset, incrementally. The results are given in Table~\ref{tab:aloi_pose}. The results show that the pose estimation error decreases as the number of training samples, $\aleph_{tr} ^o$, increases. This is due to the fact that the addition of new objects to the dataset increases the statistical representation capacity of the LHOP and the learning model of the regression algorithm. In addition, we observe that the pose estimation error observed in the experiments for $C=2$ decreases when the objects from Car category are added to a dataset of objects belonging to Cups and Cow category in the experiments with $C=3$. The performance boost is achieved by increasing the shareability of co-occurred object parts in different categories. For instance, the parts that construct the rectangular silhouettes of cows and cars can be shared in the construction of object representations in the LHOP (see Fig.~\ref{fig:aloi_examples}.

\begin{table*}[htbp]
  \centering
  \caption{Category-wise Pose estimation errors ($\epsilon$) of SVR-HOG/SVR-LHOP/L-HOG/L-LHOP/Proposed Approach for different number of categories ($C$) and training samples ($\aleph_{tr} ^o$) selected from ALOI dataset.}
	\begin{tabular}{|c|c|c|c|c|}
    \toprule
\textbf{$\aleph_{tr} ^o$} & \textbf{C=2}& \textbf{C=3} & \textbf{C=4} & \textbf{C=5} \\ 

\midrule

    \textbf{1}     & 133/103/140/97/91 & 116/99/110/97/89 & 110/95/102/95/88 & 102/94/99/95/88 \\
    \textbf{2}     & 130/100/133/95/85 & 108/93/104/88/81 & 105/91/95/88/80 & 100/94/100/91/85 \\
    \textbf{3}     & 105/91/104/86/75 & 93/83/87/83/70 & 99/86/94/84/75 & 95/81/93/75/70 \\
    \textbf{4}     & 94/86/90/73/68 & 90/79/84/73/65 & 92/77/86/72/64 & 95/75/88/71/60 \\
    \bottomrule
    \end{tabular}%
  \label{tab:aloi_pose}%
\end{table*}%

We employed two types of experiments on COIL dataset, constructing balanced and unbalanced training and test sets, in order to analyze the effect of the unbalanced data to the pose estimation performance. In the experiments, the objects are selected from Cat, Spatula, Cups and Car categories which contain $3$, $3$, $10$ and $10$ objects. Each object is rotated on a turntable by $5\degree$ from $0\degree$ to $355\degree$. 

In the experiments on balanced datasets, images of $\aleph_{tr} ^o$ number of objects are initially selected from Cat and Spatula  categories (for $C=2$), and then images of the objects selected from Cups and Car categories are incrementally added to the dataset for $C=3$ and $C=4$ category experiments. More specifically, $\aleph_{tr} ^o$ objects are randomly selected from each category and the random selection is repeated two times for each experiment. The results are shown in Table~\ref{tab:coil_pose}. We observe that the addition of new objects to the datasets decreases the pose estimation error. Moreover, we observe a remarkable performance boost when the images of the objects from the categories that have similar silhouettes, such as Cat and Cups or Spatula and Car, are used in the same dataset.

\begin{table}[htbp]
  \centering
  \caption{Category-wise Pose estimation errors ($\epsilon$) of SVR-HOG/SVR-LHOP/L-HOG/L-LHOP/Proposed Approach for different number of categories ($C$) and training samples ($\aleph_{tr} ^o$) selected from COIL dataset.}
\begin{tabular}{|c|c|c|c|c|}
    \toprule
\textbf{$\aleph_{tr} ^o$} & \textbf{C=2}& \textbf{C=3} & \textbf{C=4} \\ 
    \midrule
    1     & 125/109/120/95/85 & 120/85/103/77/68 & 110/79/95/71/62 \\ 
    2     & 120/95/114/89/77 & 93/77/81/63/59 & 104/76/92/69/51 \\
    \bottomrule
    \end{tabular}%
  \label{tab:coil_pose}%
\end{table}%

\begin{figure}[htbp]
      \centering
	  {\includegraphics[width=4.55in, height=2.95in]{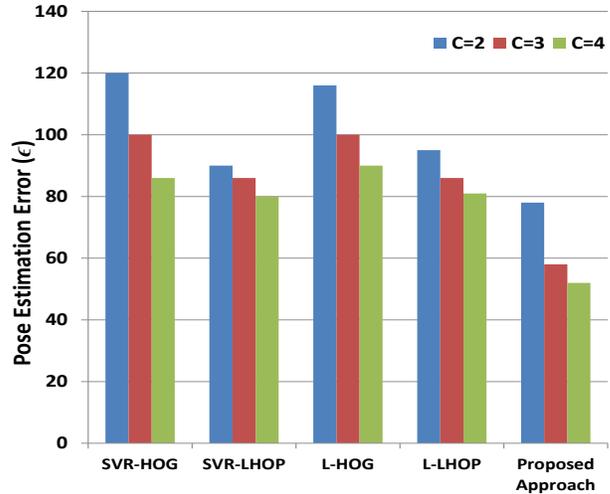}} 
      \caption{Category-wise Pose estimation errors ($\epsilon$) of the state-of-the-art algorithms and the proposed Hierarchical Compositional Approach in the experiments on COIL dataset. }

      \label{fig:coil_pose}
   \end{figure}

We prepared unbalanced datasets by randomly selecting the images of $\aleph_{te} ^o=1$ object from each category as a test sample and the images of the rest of the objects belonging to the associated category in the COIL dataset as training samples. For instance, the images of a randomly selected cat are selected as test samples and the images of the remaining two cats are selected as training samples. This procedure is repeated two times in each experiment and the average values of pose estimation errors are depicted in Fig.~\ref{fig:coil_pose}. The results show that SVR is more sensitive to the balance of the dataset and the number of training samples than the proposed approach. For instance, the difference between the pose estimation error of SVR given in Table~\ref{tab:coil_pose} and Fig.~\ref{fig:coil_pose} for $C=4$ is approximately $10\degree$, while that of the proposed Hierarchical Compositional Approach is approximately $5\degree$. 

In the next subsection, the experiments on object categorization are given.

\begin{table*}[htbp]
\caption{Categorization performance (\%) of SVM-HOG/SVM-LHOP/LR-HOG/LR-LHOP/Proposed Approach for different number of categories ($C$) and training samples ($\aleph_{tr} ^o$) selected from ALOI dataset.}
\centering
\begin{tabular}{|c|c|c|c|c|}
\toprule
\textbf{$\aleph_{tr} ^o$} & \textbf{C=2}& \textbf{C=3} & \textbf{C=4} & \textbf{C=5} \\ 
\midrule
    \textbf{1}     & 88/89/91/93/100 & 85/88/84/92/98 & 85/85/84/85/90 & 81/81/81/83/90 \\
    \textbf{2}     & 88/91/92/94/100 & 88/91/87/93/98 & 87/87/86/88/92 & 81/83/81/84/91 \\
    \textbf{3}    & 95/98/94/98/100 & 91/93/91/95/99 & 90/90/90/91/93 & 83/85/83/88/91 \\
    \textbf{4}     & 97/98/98/99/100 & 93/96/93/97/100 & 90/91/90/91/94 & 87/91/89/95/96 \\
    \bottomrule
\end{tabular}
\label{tab:aloi_cat}
\end{table*}

\subsection{Experiments on Object Categorization}
\label{sec:exp_ca}

In the Object Categorization experiments, we use the same experimental settings that are described in Section \ref{sec:exp_cw} for Category-wise Pose Estimation. 

\begin{table}[h!]
\caption{Categorization performance (\%) of SVM-HOG/SVM-LHOP/LR-HOG/LR-LHOP/Proposed Approach using COIL dataset.}
\centering
\begin{tabular}{|c|c|c|c|c|}
\toprule
\textbf{$\aleph_{tr} ^o$} & \textbf{C=2}& \textbf{C=3} & \textbf{C=4} \\ 
\midrule
\textbf{1} & 94/93/92/95/100 & 89/88/91/91/97 & 81/79/80/81/84 \\ 
\textbf{2} & 97/97/96/97/100 & 89/91/90/93/97 & 84/86/83/87/90 \\ 
\bottomrule
\end{tabular}
\label{tab:coil_cat}
\end{table}

The results of the experiments employed on ALOI dataset and balanced subsets of COIL dataset are given in Table~\ref{tab:aloi_cat} and Table~\ref{tab:coil_cat}, respectively. In these experiments, we observe that the categorization performance decreases as the number of categories increases. However, we observe that the pose estimation error decreases as the number of categories increases in the previous sections. The reason of the observation of this error difference is that the objects rotated on a turn table may provide similar silhouettes although they may belong to different categories. Therefore, addition of the images of new objects that belong to different categories may boost pose estimation performance. On the other hand, addition of the images of these new objects may decrease the categorization performance if the parts of the object cannot be shared across different categories and increase the data complexity of the feature space.

\section{Conclusion}
\label{sec:conc}

In this paper, we have proposed a compositional hierarchical approach for joint object pose estimation and categorization using a generative-discriminative learning method. The proposed approach first exposes information about pose and category of an object by extracting features from its realizations observed at different layers of LHOP in order to consider different levels of abstraction of information represented in the hierarchy. Next, we formulate joint object pose estimation and categorization problem as a sparse  optimization problem. Then, we solve the optimization problem by integrating the features extracted at each different layer using a distributed and parallel optimization algorithm.

We examine the proposed approach on benchmark 2-D multi-view image datasets. In the experiments, the proposed approach outperforms state-of-the-art Support Vector Machines for object categorization and Support Vector Regression algorithm for object pose estimation. In addition, we observe that shareability of object parts across different object categories and views may increase pose estimation performance. On the other hand, object categorization performance may decrease as the number of categories increases if parts of an object cannot be shared across different categories, and increase the data complexity of the feature space. The proposed approach can successfully estimate the pose of  objects which have view-specific statistical and geometric properties. On the other hand, the proposed feature extraction algorithms cannot provide information about the view-specific properties of symmetric or semi-symmetric objects, which leads to a decrease of the object pose estimation and categorization performance. Therefore, the ongoing work is directed towards alleviating the problems with symmetric or semi-symmetric objects.








\bibliographystyle{IEEEtranS}
\bibliography{IEEEexample}

\end{document}